# Using Convolutional Networks and Satellite Imagery to Identify Patterns in Urban Environments at a Large Scale

Adrian Albert[*]
Massachusetts Institute of Technology
Civil and Environmental Engineering
77 Massachusetts Ave
Cambridge, MA 02139
adalbert@mit.edu

Jasleen Kaur
Philips Lighting Research North America
2 Canal Park
Cambridge, MA 02141
jasleen.kaur1@philips.com

Marta C. González
Massachusetts Institute of Technology
Civil and Environmental Engineering
77 Massachusetts Ave
Cambridge, MA 02139
martag@mit.edu

## ABSTRACT

Urban planning applications (energy audits, investment, etc.) require an understanding of built infrastructure and its environment, i.e., both low-level, physical features (amount of vegetation, building area and geometry etc.), as well as higher-level concepts such as land use classes (which encode expert understanding of socio-economic end uses). This kind of data is expensive and labor-intensive to obtain, which limits its availability (particularly in developing countries). We analyze patterns in land use in urban neighborhoods using large-scale satellite imagery data (which is available worldwide from third-party providers) and state-of-the-art computer vision techniques based on deep convolutional neural networks. For supervision, given the limited availability of standard benchmarks for remote-sensing data, we obtain ground truth land use class labels carefully sampled from open-source surveys, in particular the Urban Atlas land classification dataset of 20 land use classes across 300 European cities. We use this data to train and compare deep architectures which have recently shown good performance on standard computer vision tasks (image classification and segmentation), including on geospatial data. Furthermore, we show that the deep representations extracted from satellite imagery of urban environments can be used to compare neighborhoods across several cities. We make our dataset available for other machine learning researchers to use for remote-sensing applications.

## CCS CONCEPTS

•**Computing methodologies** →**Computer vision**; **Neural networks**; •**Applied computing** →**Environmental sciences**;

## KEYWORDS

Satellite imagery, land use classification, convolutional networks

## 1 INTRODUCTION

Land use classification is an important input for applications ranging from urban planning, zoning and the issuing of business permits, to real-estate construction and evaluation to infrastructure development. Urban land use classification is typically based on surveys performed by trained professionals. As such, this task is labor-intensive, infrequent, slow, and costly. As a result, such data are mostly available in developed countries and big cities that have the resources and the vision necessary to collect and curate it; this information is usually not available in many poorer regions, including many developing countries [9] where it is mostly needed.

This paper builds on two recent trends that promise to make the analysis of urban environments a more democratic and inclusive task. On the one hand, recent years have seen significant improvements in satellite technology and its deployment (primarily through commercial operators), which allows to obtain high and medium-resolution imagery of most urbanized areas of the Earth with an almost daily revisit rate. On the other hand, the recent breakthroughs in computer vision methods, in particular deep learning models for image classification and object detection, now make possible to obtain a much more accurate representation of the composition built infrastructure and its environments.

Our contributions are to both the applied deep learning literature, and to the incipient study of "smart cities" using remote sensing data. We contrast state-of-the-art convolutional architectures (the VGG-16 [19] and ResNet [7] networks) to train classifiers that recognize broad land use classes from satellite imagery. We then use the features extracted from the model to perform a large-scale comparison of urban environments. For this, we construct a novel dataset for land use classification, pairing carefully sampled locations with ground truth land use class labels obtained from the Urban Atlas survey [22] with satellite imagery obtained from Google Maps's static API. Our dataset - which we have made available publicly for other researchers - covers, for now, 10 cities in Europe (chosen out of the original 300) with 10 land use classes (from the original 20). As the Urban Atlas is a widely-used, standardized dataset for land use classification, we hope that making this dataset available will encourage the development analyses and algorithms for analyzing the built infrastructure in urban environments. Moreover, given that satellite imagery is available virtually everywhere on the globe, the methods presented here allow for automated, rapid classification of urban environments that can potentially be applied to locations where survey and zoning data is not available.

*Land use* classification refers to the combination of physical land attributes and what cultural and socio-economic function land serves (which is a subjective judgement by experts) [2]. In this paper, we take the view that land use classes are just a useful discretization of a more continuous spectrum of patterns in the organization of urban environments. This viewpoint is illustrated in Figure 2: while some attributes (e.g., amount of built structures or vegetation) are directly interpretable, some others may not be. Nevertheless, these patterns influence, and are influenced by, socio-economic factors (e.g., economic activity), resource use (energy), and dynamic human behavior (e.g., mobility, building occupancy). We see the work on cheaply curating a large-scale land use classification dataset and comparing neighborhoods using deep representations that this paper puts forth as a necessary first step towards a granular understanding of urban environments in data-poor regions.

[*]Corresponding author.

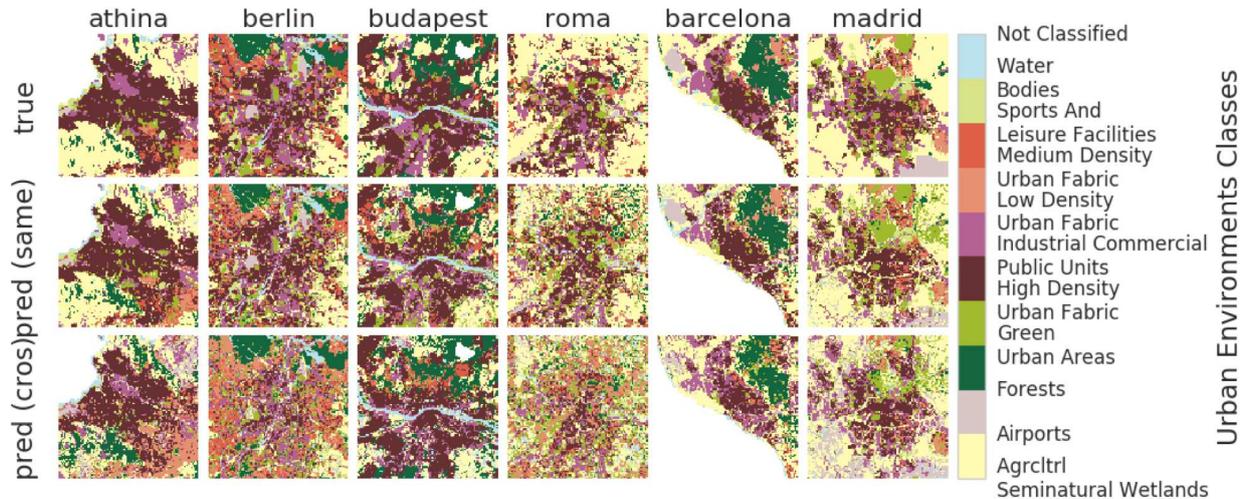

Figure 1: Urban land use maps for six example cities. We compare the ground truth (*top row*) with the predicted land use maps, either from using separate data collected from the same city (*middle row*), or using data from all other cities (*bottom row*).

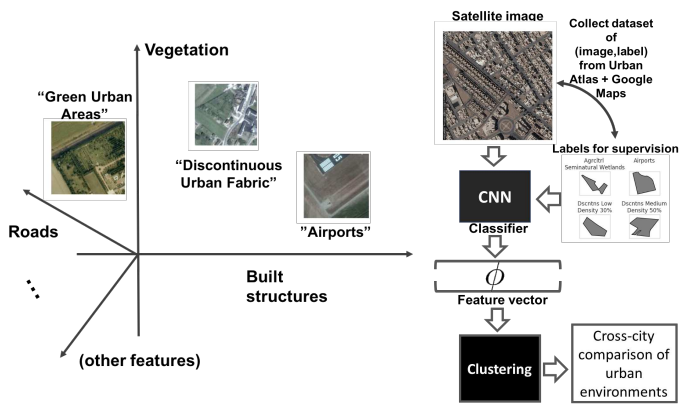

Figure 2: *Left:* Comparing urban environments via deep hierarchical representations of satellite image samples. *Right:* approach outline - data collection, classification, feature extraction, clustering, validation.

Subsequently, in Section 2 we review related studies that apply deep learning methods and other machine learning techniques to problems of land use classification, object detection, and image segmentation in aerial imagery. In Section 3 we describe the dataset we curated based on the Urban Atlas survey. Section 4 reviews the deep learning architectures we used. Section 5 describes model validation and analysis results. We conclude in Section 6.

All the code used to acquire, process, and analyze the data, as well as to train the models discussed in this paper is available at http://www.github.com/adrianalbert/urban-environments.

## 2 LITERATURE

The literature on the use of remote sensing data for applications in land use cover, urban planning, environmental science, and others, has a long and rich history. This paper however is concerned more narrowly with newer work that employs widely-available data and machine learning models - and in particular deep learning architectures - to study urban environments.

Deep learning methods have only recently started to be deployed to the analysis of satellite imagery. As such, land use classification using these tools is still a very incipient literature. Probably the first studies (yet currently only 1-2 years old) include the application of convolutional neural networks to land use classification [2] using the UC Merced land use dataset [25] (of 2100 images spanning 21 classes) and the classification of agricultural images of coffee plantations [17]. Similar early studies on land use classification that employ deep learning techniques are [21], [18], and [15]. In [11], a spatial pyramid pooling technique is employed for land use classification using satellite imagery. The authors of these studies adapted architectures pre-trained to recognize natural images from the ImageNet dataset (such as the VGG16 [19], which we also use), and fine-tuned them on their (much smaller) land use data. More recent studies use the DeepSat land use benchmark dataset [1], which we also use and describe in more detail in Section 2.1. Another topic that is closely related to ours is remote-sensing image segmentation and object detection, where modern deep learning models have also started to be applied. Some of the earliest work that develops and applies deep neural networks for this tasks is that of [13]. Examples of recent studies include [26] and [12], where the authors propose a semantic image segmentation technique combining texture features and boundary detection in an end-to-end trainable architecture.

Remote-sensing data and deep learning methods have been put to use to other related ends, e.g., geo-localization of ground-level photos via satellite images [3, 24] or predicting ground-level scene images from corresponding aerial imagery [27]. Other applications have included predicting survey estimates on poverty levels in several countries in Africa by first learning to predict levels of night lights (considered as proxies of economic activity and measured by satellites) from day-time, visual-range imagery from Google Maps, then transferring the learning from this latter task to the former [9]. Our work takes a similar approach, in that we aim to use remote-sensing data (which is widely-available for most parts of the world) to infer land use types in those locations where ground truth surveys are not available.

Urban environments have been analyzed using other types of imagery data that have become recently available. In [4, 14], the authors propose to use the same type of imagery from Google Street View to measure the relationship between urban appearance and quality of life measures such as perceived safety. For this, they hand-craft standard image features widely used in the computer vision community, and train a shallow machine learning classifier (a support vector machine). In a similar fashion, [5] trained a convolutional neural network on ground-level Street View imagery paired with a crowd-sourced mechanism for collecting ground truth labels to predict subjective perceptions of urban environments such as "beauty", "wealth", and "liveliness".

Land use classification has been studied with other new data sources in recent years. For example, ground-level imagery has been employed to accurately predict land use classes on an university campus [28]. Another related literature strand is work that uses mobile phone call records to extract spatial and temporal mobility patterns, which are then used to infer land use classes for several cities [6, 10, 20]. Our work builds on some of the ideas for sampling geospatial data presented there.

## 2.1 Existing land use benchmark datasets

Public benchmark data for land use classification using aerial imagery are still in relatively short supply. Presently there are two such datasets that we are aware of, discussed below.

**UC Merced.** This dataset was published in 2010 [25] and contains 2100 $256 \times 256$, $1m/px$ aerial RGB images over 21 land use classes. It is considered a "solved problem", as modern neural network based classifiers [2] have achieved $> 95\%$ accuracy on it.

**DeepSat.** The DeepSat [1] dataset[1] was released in 2015. It contains two benchmarks: the *Sat-4* data of 500, 000 images over 4 land use classes (*barren land, trees, grassland, other*), and the *Sat-6* data of 405, 000 images over 6 land use classes (*barren land, trees, grassland, roads, buildings, water bodies*). All the samples are $28 \times 28$ in size at a $1m/px$ spatial resolution and contain 4 channels (red, green, blue, and NIR - near infrared). Currently less than two years old, this dataset is already a "solved problem", with previous studies [15] (and our own experiments) achieving classification accuracies of over 99% using convolutional architectures. While useful as input for pre-training more complex models, (e.g., image segmentation), this dataset does not allow to take the further steps for detailed land use analysis and comparison of urban environments across cities, which gap we hope our dataset will address.

**Other open-source efforts.** There are several other projects that we are aware of related to land use classification using open-source data. The TerraPattern[2] project uses satellite imagery from Google Maps (just like we do) paired with truth labels over a large number (450) of detailed classes obtained using the Open Street Map API[3]. (Open Street Maps is a comprehensive, open-access, crowd-sourced mapping system.) The project's intended use is as a search tool for satellite imagery, and as such, the classes they employ are very specific, e.g., baseball diamonds, churches, or roundabouts. The authors use a ResNet architecture [7] to train a classification model, which they use to embed images in a high-dimensional feature space, where "similar" images to an input image can be identified. A second open-source project related to ours is the DeepOSM[4], in which the authors take the same approach of pairing OpenStreetMap labels with satellite imagery obtained from Google Maps, and use a convolutional architecture for classification. These are excellent starting points from a practical standpoint, allowing interested researchers to quickly familiarize themselves with programming aspects of data collection, API calls, etc.

## 3 THE URBAN ENVIRONMENTS DATASET

### 3.1 Urban Atlas: a standard in land use analysis

The Urban Atlas [22] is an open-source, standardized land use dataset that covers ∼ 300 European cities of 100, 000 inhabitants or more, distributed relatively evenly across major geographical and geopolitical regions. The dataset was created between 2005-2011 as part of a major effort by the European Union to provide a uniform framework for the geospatial analysis of urban areas in Europe. Land use classification is encoded via detailed polygons organized in commonly-used GIS/ESRI shape files. The dataset covers 20 standardized land use classes. In this work we selected classes of interest and consolidated them into 10 final classes used for analysis (see Figure 3). Producing the original Urban Atlas dataset required fusing several data sources: high and medium-resolution satellite imagery, topographic maps, navigation and road layout data, and local zoning (cadastral) databases. More information on the methodology used by the Urban Atlas researchers can be obtained from the European Environment Agency[5]. We chose expressly to use the Urban Atlas dataset over other sources (described in Section 2.1 because *i)* it is a comprehensive and consistent survey at a large scale, which has been extensively curated by experts and used in research, planning, and socio-economic work over the past decade, and *ii)* the land use classes reflect higher-level (socio-economic, cultural) functions of the land as used in applications.

We note that there is a wide variance in the distribution of land use classes across and within the 300 cities. Figure 3 illustrates the differences in the distribution in ground truth polygon areas for each of the classes for three example cities (Budapest, Rome, Barcelona) from the dataset (from Eastern, Central, and Western Europe, respectively). This wide disparity in the spatial distribution patterns of different land use classes and across different cities motivates us to design a careful sampling procedure for collecting training data, described in detail below.

---

[1]Available at http://csc.lsu.edu/~saikat/deepsat/.

[2]http://www.terrapattern.com/
[3]http://www.openstreetmap.org
[4]https://github.com/trailbehind/DeepOSM
[5]http://www.eea.europa.eu/data-and-maps/data/urban-atlas/

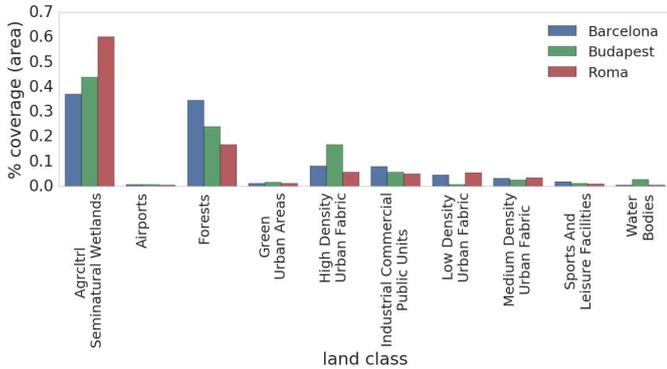

**Figure 3: Ground truth land use distribution (by area) for three example cities in the Urban Environments dataset.**

## 3.2 Data sampling and acquisition

We set out to develop a strategy to obtain high-quality samples of the type (satellite image, ground truth label) to use in training convolutional architectures for image classification. Our first requirement is to do this solely with freely-available data sources, as to keep costs very low or close to zero. For this, we chose to use the Google Maps Static API[6] as a source of satellite imagery. This service allows for 25,000 API requests/day free of charge. For a given sampling location given by (latitude, longitude), we obtained $224 \times 224$ images at a zoom level 17 (around $1.20m/px$ spatial resolution, or $\sim 250m \times 250m$ coverage for an image).

The goals of our sampling strategy are twofold. First, we want to ensure that the resulting dataset is as much as possible balanced with respect to the land use classes. The challenge is that the classes are highly imbalanced among the ground truth polygons in the dataset (e.g., many more polygons are agricultural land and isolated structures than airports). Second, the satellite images should be representative of the ground truth class associated to them. To this end, we require that the image contain at least 25% (by area) of the associated ground truth polygon. Thus, our strategy to obtain training samples is as follows (for a given city):

- Sort ground truth polygons in decreasing order according to their size, and retain only those polygons with areas larger than $\frac{1}{4}(224 \times 1.2m)^2 = 0.06 km^2$;
- From each decile of the distribution of areas, sample a proportionally larger number of polygons, such that some of the smaller polygons also are picked, and more of the larger ones;
- For each picked polygon, sample a number of images proportional to the area of the polygon, and assign each image the polygon class as ground truth label;

---

[6] https://developers.google.com/maps/documentation/static-maps/

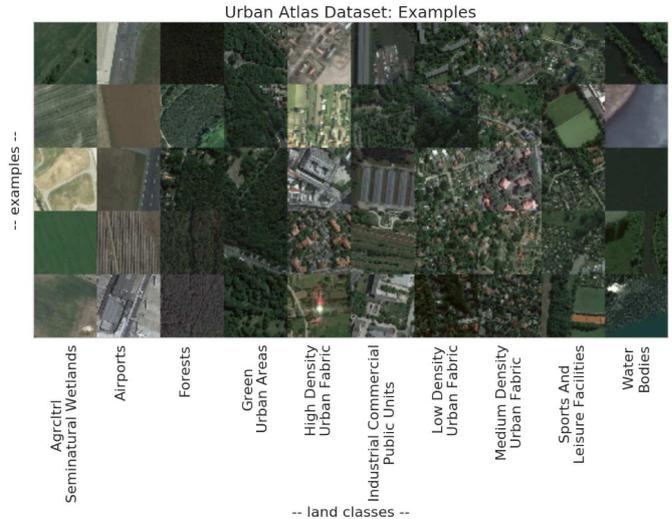

**Figure 4: Example satellite images for the original land use classes in the Urban Atlas dataset.**

Example satellite images for each of the 10 land use classes in the Urban Environments dataset are given in Figure 4. Note the significant variety (in color schemes, textures, etc) in environments denoted as having the same land use class. This is because of several factors, including the time of the year when the image was acquired (e.g., agricultural lands appear different in the spring than in the fall), the different physical form and appearance of environments that serve the same socioeconomic or cultural function (e.g., green urban areas may look very different in different cities or in even in different parts of the same city; what counts as "dense urban fabric" in one city may not be dense at all in other cities), and change in the landscape during the several years that have passed since the compilation of the Urban Atlas dataset and the time of acquisition of the satellite image (e.g., construction sites may not reflect accurately anymore the reality on the ground).

Apart from these training images, we constructed ground truth rasters to validate model output for each city. For that, we defined uniform validation grids of $100 \times 100$ ($25km \times 25km$) around the (geographical) center of a given city of interest. We take a satellite image sample in each grid cell, and assign to it as label the class of the polygon that has the maximum intersection area with that cell. Examples of land use maps for the six cities we analyze here are given in Figure 1 (top row). There, each grid cell is assigned the class of the ground truth polygon whose intersection with the cell has maximum coverage fraction by area. Classes are color-coded following the original Urban Atlas documentation.

In Table 1 we present summaries of the training (left) and validation (right) datasets we used for the analysis in this paper. The validation dataset consists of the images sampled at the centers of each cell in the $25km \times 25km$ grid as discussed above. This dataset consists of $\sim 140,000$ images distributed across 10 urban environment classes from 6 cities: Roma (Rome), Madrid, Berlin, Budapest, Barcelona, and Athina (Athens). Because of the high variation in appearance upon visual inspection, we chose to consolidate several classes from the original dataset, in particular classes that indicated urban fabric into "High Density Urban Fabric", "Medium Density Urban Fabric, and "Low Density Urban Fabric". As mentioned above and illustrated in Figure 3, we did notice a great disparity in the numbers and distribution of ground truth polygons for other example cities that we investigated in the Urban Atlas dataset. As such,for the analysis in this paper, we have chosen cities where enough ground truth polygons were available for each class (that is, at least 50 samples) to allow for statistical comparisons.

## 4 EXPERIMENTAL SETUP
### 4.1 Neural network architectures and training
For all experiments in this paper we compared the VGG-16 [19] and ResNet [7, 8] architectures.

**VGG-16.** This architecture [19] has become one of the most popular models in computer vision for classification and segmentation tasks. It consists of 16 trainable layers organized in blocks. It starts with a 5-block convolutional base of neurons with $3 \times 3$ receptive fields (alternated with max-pooling layers that effectively increase the receptive field of neurons further downstream). Following each convolutional layer is a ReLU activation function [19]. The feature maps thus obtained are fed into a set of fully-connected layers (a deep neural network classifier). See Table 2 for a summary.

**ResNet.** This architecture [7, 8] has achieved state-of-the-art performance on image classification on several popular natural image benchmark datasets. It consists of blocks of convolutional layers, each of which is followed by a ReLU non-linearity. As before, each block in the convolutional base is followed by a max-pooling operation. Finally, the output of the last convolutional layer serves as input feature map for a fully-connected layer with a softmax activation function. The key difference in this architecture is that *shortcut* connections are implemented that skip blocks of convolutional layers, allowing the network to learn residual mappings between layer input and output. Here we used an implementation with 50 trainable layers per [7]. See Table 3 for a summary.

**Transfer learning.** As it is common practice in the literature, we have experimented with training our models on the problem of interest (urban environment classification) starting from architectures pre-trained on datasets from other domains (*transfer learning*). This procedure has been shown to yield both better performance and faster training times, as the network already has learned to recognize basic shapes and patterns that are characteristic of images across many domains (e.g., [9, 12, 15]). We have implemented the following approaches: *1)* we used models pre-trained on the ImageNet dataset, then further fine-tuned them on the Urban Atlas dataset; and *2)* we pre-trained on the DeepSat dataset (See Section 2), then further refined on the Urban Atlas dataset. As expected, the latter strategy - first training a model (itself pre-trained on ImageNet data) on the DeepSat benchmark, and the further refining on the Urban Atlas dataset - yielded the best results, achieving increases of around 5% accuracy for a given training time.

Given the large amount of variation in the visual appearance of urban environments across different cities (because of different climates, different architecture styles, various other socio-economic factors), it is of interest to study to what extent a model learned on one geographical location can be applied to a different geographical location. As such, we perform experiments in which we train a model for one (or more) cities, then apply the model to a different set of cities. Intuitively, one would expect that, the more neighborhoods and other urban features at one location are similar to those at a different location, the better learning would transfer, and the higher the classification accuracy obtained would be. Results for these experiments are summarized in Figure 6.

### 4.2 Comparing urban environments
We next used the convolutional architectures to extract features for validation images. As in other recent studies (e.g., [9]), we use the last layer of a network as feature extractor. This amounts to feature vectors of $D = 4096$ dimensions for the VGG16 architecture and $D = 2048$ dimensions for the ResNet-50 architecture. The codes $x \in \mathbb{R}^D$ are the image representations that either network derives as most representative to discriminate the high-level land use concepts it is trained to predict.

We would like to study how "similar" different classes of urban environments are across two example cities (here we picked Berlin and Barcelona, which are fairly different from a cultural and architectural standpoint). For this, we focus only on the $25km \times 25km$, $100 \times 100$-cell grids around the city center as in Figure 1. To be able to quantify similarity in local urban environments, we construct a KD-tree $\mathcal{T}$ (using a high-performance implementation available in the Python package `scikit-learn` [16]) using all the gridded samples. This data structure allows to find $k$-nearest neighbors of a query image in an efficient way. In this way, the feature space can be probed in an efficient way.

## 5 RESULTS AND DISCUSSION
In Figure 1 we show model performance on the $100 \times 100$ ($25km \times 25km$) raster grids we used for testing. The top row shows ground truth grids, where the class in each cell was assigned as the most prevalent land use class by area (see also Section 3). The bottom row shows model predictions, where each cell in a raster is painted in the color corresponding to the maximum probability class estimated by the model (here ResNet-50). Columns in the figure show results for each of the 6 cities we used in our dataset. Even at a first visual inspection, the model is able to recreate from satellite imagery qualitatively the urban land use classification map.

Further, looking at the individual classes separately and the confidence of the model in its predictions (the probability distribution over classes computed by the model), the picture is again qualitatively very encouraging. In Figure 5 we show grayscale raster maps encoding the spatial layout of the class probability distribution for one example city, Barcelona. Particularly good qualitative agreement is observed for agricultural lands, water bodies, industrial, public, and commercial land, forests, green urban areas, low density urban fabric, airports, and sports and leisure facilities. The model appears to struggle with reconstructing the spatial distribution of roads, which is not unexpected, given that roads typically appear in many other scenes that have a different functional classification for urban planning purposes.

Table 1: Urban Environments dataset: sample size summary.

(a) Dataset used for training & validation (80% and 20%, respectively)

| class/city | athina | barcelona | berlin | budapest | madrid | roma | class total |
|---|---|---|---|---|---|---|---|
| Agricultural + Semi-natural areas + Wetlands | 4347 | 2987 | 7602 | 2211 | 4662 | 4043 | 25852 |
| Airports | 382 | 452 | 232 | 138 | 124 | 142 | 1470 |
| Forests | 1806 | 2438 | 7397 | 1550 | 2685 | 2057 | 17933 |
| Green urban areas | 990 | 722 | 1840 | 1342 | 1243 | 1401 | 7538 |
| High Density Urban Fabric | 967 | 996 | 8975 | 6993 | 2533 | 3103 | 23567 |
| Industrial, commercial, public, military and pr... | 1887 | 2116 | 4761 | 1850 | 3203 | 2334 | 16151 |
| Low Density Urban Fabric | 1424 | 1520 | 2144 | 575 | 2794 | 3689 | 12146 |
| Medium Density Urban Fabric | 2144 | 1128 | 6124 | 1661 | 1833 | 2100 | 14990 |
| Sports and leisure facilities | 750 | 1185 | 2268 | 1305 | 1397 | 1336 | 8241 |
| Water bodies | 537 | 408 | 1919 | 807 | 805 | 619 | 5095 |
| city total | 15234 | 13952 | 43262 | 18432 | 21279 | 20824 | 132983 |

(b) $25km \times 25km$ ground truth test grids (fractions of city total)

| class / city | athina | barcelona | berlin | budapest | madrid | roma |
|---|---|---|---|---|---|---|
| Agricultural + Semi-natural areas + Wetlands | 0.350 | 0.261 | 0.106 | 0.181 | 0.395 | 0.473 |
| Airports | 0.003 | 0.030 | 0.013 | 0.000 | 0.044 | 0.006 |
| Forests | 0.031 | 0.192 | 0.087 | 0.211 | 0.013 | 0.019 |
| Green urban areas | 0.038 | 0.030 | 0.072 | 0.027 | 0.125 | 0.054 |
| High Density Urban Fabric | 0.389 | 0.217 | 0.284 | 0.365 | 0.170 | 0.215 |
| Industrial, commercial, public, military and pr... | 0.109 | 0.160 | 0.190 | 0.096 | 0.138 | 0.129 |
| Low Density Urban Fabric | 0.016 | 0.044 | 0.012 | 0.006 | 0.036 | 0.029 |
| Medium Density Urban Fabric | 0.041 | 0.025 | 0.129 | 0.045 | 0.042 | 0.047 |
| Sports and leisure facilities | 0.017 | 0.034 | 0.080 | 0.025 | 0.036 | 0.025 |
| Water bodies | 0.005 | 0.006 | 0.026 | 0.044 | <0.001 | 0.004 |

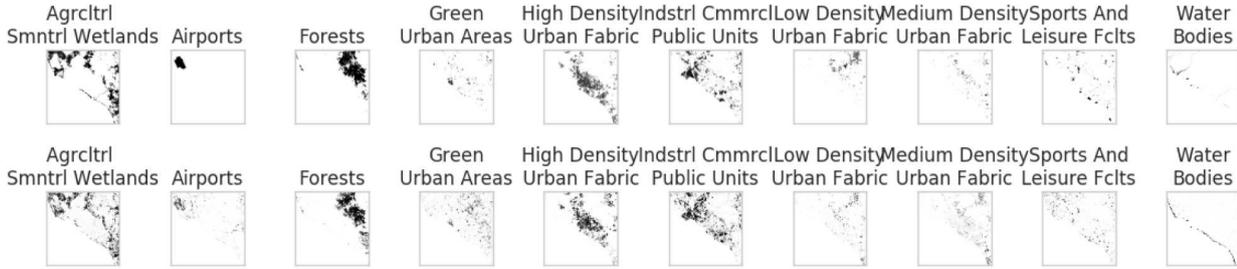

Figure 5: Barcelona: ground truth (*top*) and predicted probabilities (*bottom*).

Table 2: The VGG16 architecture [19].

| Block 1 | Block 2 | Block 3 | Block 4 | Block 5 | Block 6 |
|---|---|---|---|---|---|
| Conv(3,64) | Conv(3,128) | Conv(3,256) | Conv(3,512) | Conv(3,512) | FC(4096) |
| Conv(3,64) | Conv(3,128) | Conv(3,256) | Conv(3,512) | Conv(3,512) | FC(4096) |
| Max-Pool(2,2) | Max-Pool(2,2) | Conv(3,256) | Conv(3,512) | Conv(3,512) | FC($N_{\text{classes}}$) |
| | | Max-Pool(2,2) | Max-Pool(2,2) | Max-Pool(2,2) | SoftMax |

Table 3: The ResNet-50 architecture [7].

| Block 1 | Block 2 | Block 3 | Block 4 | Block 5 | Block 6 |
|---|---|---|---|---|---|
| Conv(7,64) | 3x[Conv(1,64) | 4x[Conv(1,128) | 6x[Conv(1,256) | 3x[Conv(1,512) | FC($N_{\text{classes}}$) |
| Max-Pool(3,2) | Conv(3,64) | Conv(3,128) | Conv(3,256) | Conv(3,512) | SoftMax |
| | Conv(3,256)] | Conv(1,512)] | Conv(1,1024)] | Conv(1,2048)] | |

## 5.1 Classification results

We performed experiments training the two architectures described in Section 4 on datasets for each of the 6 cities considered, and for a combined dataset (all) of all the cities. The diagonal in Figure 6 summarizes the (validation set) classification performance for each model. All figures are averages computed over balanced subsets of 2000 samples each. While accuracies or $\sim 0.70 - 0.80$ may not look as impressive as those obtained by convolutional architectures on well-studied benchmarks and other classification tasks (e.g., natural images from ImageNet or small aerial patches from Deep-Sat), this only attests to the difficulty of the task of understanding high-level, subjective concepts of urban planning in complex urban environments. First, satellite imagery typically contains much more semantic variation than natural images (as also noted, e.g., in [2, 13]), i.e., there is no "central" concept that the image is of (unlike the image of a cat or a flower). Second, the type of labels we use for supervision are higher-level concepts (such as "low density urban fabric", or "sports and leisure facilities"), which are much less specific than more physical land features e.g., "buildings" or "trees" (which are classes used in the DeepSat dataset). Moreover, top-down imagery poses specific challenges to convolutional architectures, as these models are inherently not rotationally-symmetric. Urban environments, especially from from a top-down point of view, come in many complex layouts, for which rotations are irrelevant. Nevertheless, these results are encouraging, especially since this is a harder problem by focusing on wider-area images and on higher-level, subjective concepts used in urban planning rather than on the standard, lower-level physical features such as in [1] or [17]. This suggests that such models may be useful feature extractors. Moreover, as more researchers tackle problems with the aid of satellite imagery (which is still a relatively under-researched source of data in the machine learning community), more open-source datasets (like this one) are released, performance will certainly improve. For the remainder of this section we report results using the ResNet-50 architecture [7], as it consistently yielded (if only slightly) better classification results in our experiments than the VGG-16 architecture.

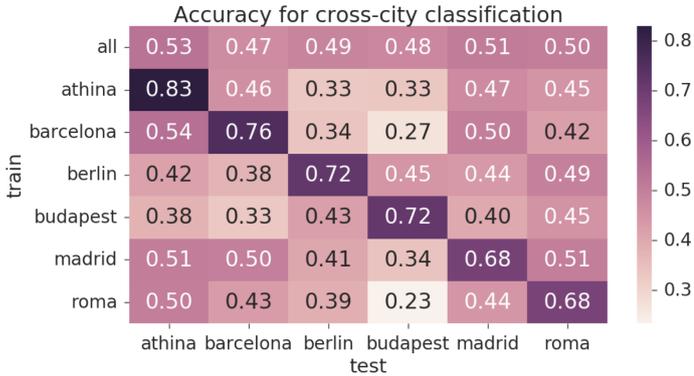

**Figure 6: Transferability (classification accuracy) of models learned at one location and applied at another. Training on a more diverse set of cities (`all`) yields encouraging results compared with just pairwise training/testing.**

**Transfer learning and classification performance.** Next, we investigated how models trained in one setting (city or set of cities) perform when applied to other geographical locations. Figure 6 summarizes these experiments. In general, performance is poor when training on samples from a given city and testing on samples from a different city (the off-diagonal terms). This is expected, as these environments can be very different in appearance for cities as different as e.g., Budapest and Barcelona. However, we notice that a more diverse set (`all`) yields better performance when applied at different locations than models trained on individual cities. This is encouraging for our purpose of analyzing the high level "similarity" of urban neighborhoods via satellite imagery.

We next looked at per-class model performance to understand what types of environments are harder for the model to distinguish. Figure 7 shows such an example analysis for three example cities, of which a pair is "similar" according to Figure 6 (Rome and Barcelona), and another dissimilar (Rome and Budapest). The left panel shows model performance when training on samples from Barcelona, and predicting on test samples from Barcelona (intra-city). The middle panel shows training on Rome, and predicting on test samples in Barcelona, which can be assumed to be "similar" to Rome from a cultural and architectural standpoint (both Latin cities in warm climates). The right figure shows training on Barcelona, and predicting on test samples in Budapest, which can be assumed a rather different city from a cultural and architectural standpoint. For all cases, the classes that the model most struggles with are "High Density Urban Fabric", "Low Density Urban Fabric, and "Medium Density Urban Fabric". Considerable overlap can be noticed between these classes - which is not surprising given the highly subjective nature of these concepts. Other examples where the model performance is lower is forests and low-density urban areas being sometimes misclassifed as "green urban areas", which, again, is not surprising. This is especially apparent in the cross-city case, where the model struggles with telling apart these classes. For both the case of training and testing on "different cities" (Budapest and Barcelona) and on "similar" cities (Rome and Barcelona), we note that airports and forests are relatively easier to distinguish. However, more subjective, high-level urban-planning concepts such as "high density urban fabric" are harder to infer (and more easily confused with "medium density" or "low density" urban fabric) in the case of more similar cities (Rome and Barcelona) rather than dissimilar cities (Budapest and Barcelona). Urban environments containing sports and leisure facilities and green areas are under this view more similar between Rome and Barcelona than they are between Budapest and Barcelona.

**Choosing the spatial scale: sensitivity analysis.** So far, we have presented results assuming that tiles of $250m$ is an appropriate spatial scale for this analysis. Our intuition suggested that tiles of this size have enough variation and information to be recognized (even by humans) as belonging to one of the high-level concepts of land use classes that we study in this paper. However, one can find arguments in favor of smaller tile sizes, e.g., in many cities the size of a typical city block is $100m$. Thus, we trained models at different spatial scales and computed test-set accuracy values for three example cities, Barcelona, Roma, and Budapest - see Figure 8. It is apparent that, for all example cities, smaller spatial scales ($50m$, $100m$, $150m$) that we analyzed yield poorer performance than the scale we chose for the analysis in this paper ($250m$). This is likely because images at smaller scales do not capture enough variation in urban form (number and type of buildings, relative amount of vegetation, roads etc.) to allow for discriminating between concepts that are fairly high-level. This is in contrast with a benchmark such as DeepSat [1] that focuses on lower-level, physical concepts ("trees", "buildings", etc.). There, a good spatial scale is by necessity smaller ($28m$ for DeepSat), as variation in appearance and compositional elements is unwanted.

## 5.2 Comparing urban environments

Finally, we set to understand, at least on an initial qualitative level, how "similar" urban environments are to one another, across formal land use classes and geographies. Our first experiment was to project sample images for each class and city in this analysis to lower-dimensional manifolds, using the `t-SNE` algorithm [23]. This serves the purpose of both visualization (as `t-SNE` is widely used for visualizing high-dimensional data), as well as for providing an initial, coarse *continuous representation* of urban land use classes. In our experiments, we used balanced samples of size $N = 6000$, or 100 samples for each of the 10 classes for each city. We extracted features for each of these samples using the `all` models (trained on a train set with samples across all cities except for the test one).

Figure 9 visualizes such `t-SNE` embeddings for the six cities in our analysis. For most cities, classes such as low density urban fabric, forests, and water bodies are well-resolved, while sports and leisure facilities seem to consistently blend into other types of environments (which is not surprising, given that these types of facilities can be found within many types of locations that have a different formal urban planning class assigned). Intriguing differences emerge in this picture among the cities. For example, green urban spaces seem fairly well resolved for most cities. Commercial neighborhoods in Barcelona seem more integrated with the other types of environments in the city, whereas for Berlin they appear more distinct. Urban water bodies are more embedded with urban parks for Barcelona than for other cities. Such reasoning (with more rigorous quantitative analysis) can serve as coarse way to benchmark and compare neighborhoods as input to further analysis about e.g., energy use, livelihood, or traffic in urban environments.

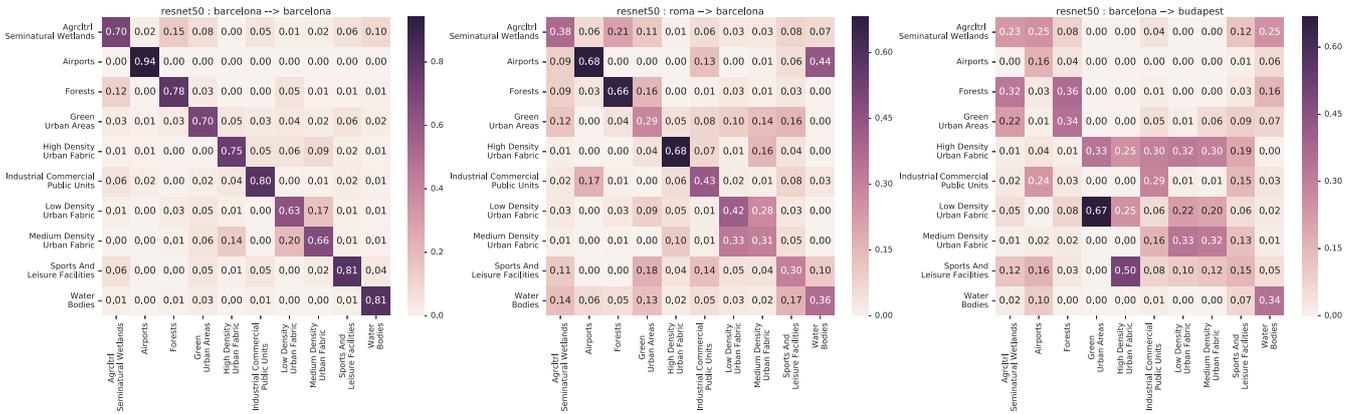

Figure 7: Example classification confusion matrix for land use inference. *Left*: training and testing on Barcelona; *Middle*: training on Rome, testing on Barcelona; *Right:* training on Rome, predicting on Budapest.

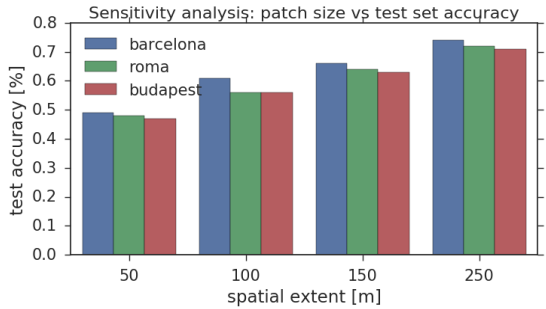

Figure 8: Sensitivity of training patch size vs test accuracy.

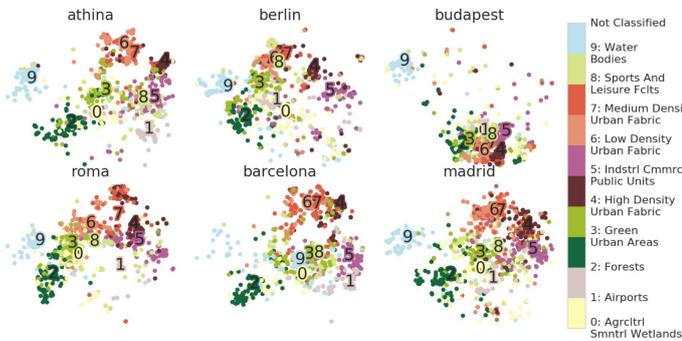

Figure 9: `t-SNE` visualization (the first 2 dimensions) of urban environments (satellite image samples) across six cities.

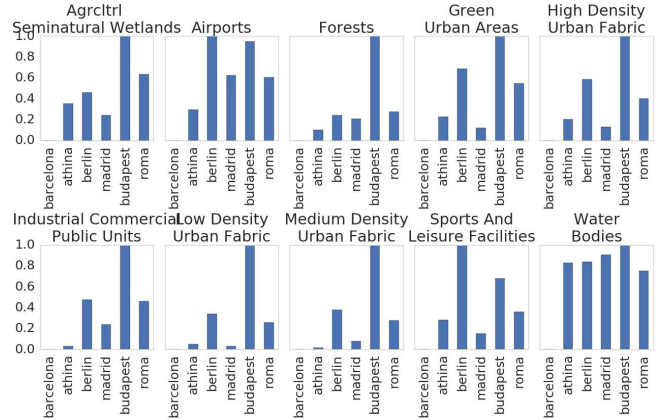

Figure 10: Comparing urban environments across cities (with reference to Barcelona) We show relative inter-city similarity measures computed as the sum of squares across the clusters in Figure 9.

We further illustrate how "similar" the six cities we used throughout this analysis are starting off the embeddings plots in Figure 9. For each land use class, we compute intra-city sum of squares in the 2-d t-SNE embedding, and display the results in Figure 10. Note that the distances are always shown with Barcelona as a reference point (chosen arbitrarily). For each panel, the normalization is with respect to the largest inter-city distance for that land use class. This visualization aids quick understanding of similarity between urban environments. For example, agricultural lands in Barcelona are most dissimilar to those in Budapest. Airports in Barcelona are most similar to those in Athens, and most dissimilar to those in Berlin and Budapest. Barcelona's forests and parks are most dissimilar to Budapest's. Water bodies in Barcelona are very dissimilar to all other cities. This point is enforced by Figure 11 below, which suggests that areas marked as water bodies in Barcelona are ocean waterfronts, whereas this class for all other cities represents rivers or lakes.

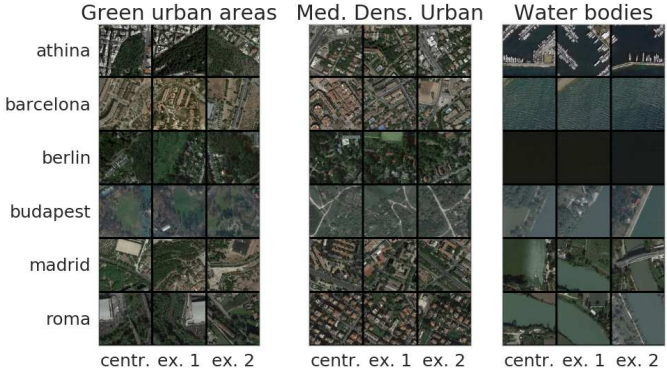

**Figure 11: Samples from three urban environments across our 6 example cities. We sampled the 2-d t-SNE embedding of Figure 9 and queried for the closest real sample to the centroid using an efficient KD-tree search.**

Finally, we explore the feature maps extracted by the convolutional model in order to illustrate how "similar" the six cities we used throughout this analysis are across three example environments, green urban areas, water bodies, and medium density urban fabric. For each city and land use class, we start off the centroid of the point cloud in the 2-d space of Figure 9, and find the nearest several samples using the KD-tree method described in Section 4. We present the results in Figure 11. Visual inspection indicates that the model has learned useful feature maps about urban environments: the sample image patches show a very good qualitative agreement with the region of the space where they're sampled from, indicated by the land use class of neighboring points. Qualitatively, it is clear that the features extracted from the top layer of the convolutional model allow a comparison between urban environments by high-level concepts used in urban planning.

## 6 CONCLUSIONS

This paper has investigated the use of convolutional neural networks for analyzing urban environments through satellite imagery at the scale of entire cities. Given the current relative dearth of *labeled* satellite imagery in the machine learning community, we have constructed an open dataset of over 140, 000 samples over 10 consistent land use classes from 6 cities in Europe. As we continue to improve, curate, and expand this dataset, we hope that it can help other researchers in machine learning, smart cities, urban planning, and related fields in their work on understanding cities.

We set out to study similarity and variability across urban environments, as being able to quantify such patterns will enable richer applications in topics such as urban energy analysis, infrastructure benchmarking, and socio-economic composition of communities. We formulated this as a two-step task: first predicting urban land use classes from satellite imagery, then turning this (rigid) classification into a continuous spectrum by embedding the features extracted from the convolutional classifier into a lower-dimensional manifold. We show that the classification task achieves encouraging results, given the large variety in physical appearance of urban environments having the same functional class. Moreover, we exemplify how the features extracted from the convolutional network allow for identifying "neighbors" of any given query image, allowing a rich comparison analysis of urban environments by their visual composition.

The analysis in this paper shows that some types urban environments are easier to infer than others, both in the intra- and inter-city cases. For example, in all our experiments, the models had most trouble telling apart "high", "medium", and "low" density urban environments, attesting to the subjectivity of such a high-level classification for urban planning purposes. However, agricultural lands, forests, and airports tend to be visually similar across different cities - and the amount of relative dissimilarity can be quantified using the methods in this paper. Green urban areas (parks) are generally similar to forests or to leisure facilities, and the models do better in the intra-city case than predicting across cities. How industrial areas look is again less geography-specific: inter-city similarity is consistently larger than intra-city similarity. As such, for several classes we can expect learning to transfer from one geography to another. Thus, while it is not news that some cities are more "similar" than others (Barcelona is visually closer to Athens than it is to Berlin), the methodology in this paper allows for a more quantitative and practical comparison of similarity.

By leveraging satellite data (available virtually world-wide), this approach may allow for a low-cost way to analyze urban environments in locations where ground truth information on urban planning is not available. As future directions of this work, we plan to *i)* continue to develop more rigorous ways to compare and benchmark urban neighborhoods, going deeper to physical elements (vegetation, buildings, roads etc.); *ii)* improve and further curate the open Urban Environments dataset; and *iii)* extend this type of analysis to more cities across other geographical locations.

## A  PRACTICAL TRAINING DETAILS.

We split our training data into a training set (80% of the data) and a validation set (the remaining 20%). This is separate from the data sampled for the ground truth raster grids for each city, which we only used at test time. We implemented the architectures in the open-source deep learning framework Keras[7](with a TensorFlow[8] backend). In all our experiments, we used popular data augmentation techniques, including random horizontal and vertical flipping of the input images, random shearing (up to 0.1 radians), random scaling (up to 120%), random rotations (by at most 15 degrees either direction). Input images were $224 \times 224 \times 3$ pixels in size (RGB bands). For all experiments, we used stochastic gradient descent (with its Adadelta variant) to optimize the network loss function (a standard multi-class cross-entropy), starting with a learning rate of 0.1, and halving the rate each 10 epochs. We trained our networks for at most 100 epochs, with 2000 samples in each epoch, stopping the learning process when the accuracy on the validation set did not improve for more than 10 epochs. Given the inherent imbalance of the classes, we explicitly enforced that the minibatches used for training were relatively balanced by a weighted sampling procedure. For training, we used a cluster of 4 NVIDIA K80 GPUs, and tested our models on a cluster of 48 CPUs.

---
[7]https://github.com/fchollet/keras
[8]www.tensorflow.org